\documentclass{jdmdh}
\usepackage{array}
\usepackage{pgfplots}

\usepackage{graphicx}
\usepackage{tabularx}
\usepackage{multirow}

\usepackage[T1]{fontenc}
\usepackage[utf8]{inputenc}

\usepackage{amsfonts}
\usepackage{amsmath} 

\usepackage[linesnumbered,ruled,vlined]{algorithm2e}

\usepackage{float}

\usepackage{caption}

\renewcommand{\refname}{References\\}

\title{Processing the Structure of Documents: Logical Layout Analysis of Historical Newspapers in French}

\author[1]{Nicolas Gutehrlé}
\author[1, 2]{Iana Atanassova}
\affil[1]{Centre de Recherches Interdisciplinaires et Transculturelles (CRIT) \protect\\
Université de Bourgogne Franche-Comté, \protect\\
30 rue Mégevand, 25000, Besançon, France\\
} 
\affil[2]{Institut Universitaire de France (IUF)}

\corrauthor{Nicolas Gutehrlé}{nicolas.gutehrle@univ-fcomte.fr}

\begin{document}

\maketitle

\abstract{
\noindent
\textbf{Background.} In recent years, libraries and archives led important digitisation campaigns that opened the access to vast collections of historical documents. While such documents are often available as XML ALTO documents, they lack information about their logical structure. 
In this paper, we address the problem of Logical Layout Analysis applied to historical documents in French. We propose a rule-based method, that we evaluate and compare with two Machine-Learning models, namely RIPPER and Gradient Boosting. Our data set contains French newspapers, periodicals and magazines, published in the first half of the twentieth century in the Franche-Comté Region. 

\noindent
\textbf{Results.} Our rule-based system outperforms the two other models in nearly all evaluations. It has especially better Recall results, indicating that our system covers more types of every logical label than the other two models. When comparing RIPPER with Gradient Boosting, we can observe that Gradient Boosting has better Precision scores but RIPPER has better Recall scores.

\noindent
\textbf{Conclusions.} The evaluation shows that our system outperforms the two Machine Learning models, and provides significantly higher Recall. It also confirms that our system can be used to produce annotated data sets that are large enough to envisage Machine Learning or Deep Learning approaches for the task of Logical Layout Analysis. Combining rules and Machine Learning models into hybrid systems could potentially provide even better performances. Furthermore, as the layout in historical documents evolves rapidly, one possible solution to overcome this problem would be to apply Rule Learning algorithms to bootstrap rule sets adapted to different publication periods.

}


\section{Background\\}

\strut
\vspace{-4ex}

One important challenge in digital humanities is the efficient exploitation and processing of scanned textual documents (archives, documentary funds, ...). For example, historical documents such as newspaper archives are prime resources for historians \citep{Tibbo2007PrimarilyHI}. 
Thanks to the important digitisation campaigns led by libraries and archives, vast collections of historical documents have been made easily accessible. However, the majority of these documents are available only as scanned images (e.g. in PDF format) which makes them difficult to explore in a text processing perspective. Extracting the text content from such documents requires at least the following three steps: Optical Character Recognition (OCR), physical layout analysis (PLA) and Logical Layout Analysis (LLA).

\textit{Physical layout analysis (PLA)}, which is also sometimes called \textit{document layout analysis}, consists in identifying physical regions of the document, with their text content and boundaries. Such regions can correspond to sections and lines of text, but also to figures, tables, etc. PLA also defines the reading order of the document, which corresponds to the linear order in which the different regions appear. This is particularly important for documents that have multi-column layouts. One commonly used output format of PLA is the XML ALTO format (\url{https://www.loc.gov/standards/alto/}). \textit{Logical layout analysis (LLA)}, sometimes called \textit{logical structure derivation} and \textit{structure understanding}, consists in identifying the document structure elements and their categories i.e. title, header, paragraph, table, etc. Such logical elements can integrate one or more regions in the document that have been identified by PLA.
Physical and logical layout analyses are necessary steps in the processing of documents for a large number of applications, including information retrieval, information extraction, table of content extraction, text syntheses, and more broadly document understanding.

In this article we focus on the problem of Logical Layout Analysis (LLA). We describe a methodology for Logical Layout Analysis, where logical labels are assigned to physical layout entities. The input of our processing pipeline is the physical layout analysis of documents in the XML ALTO format. 

An important body of research around physical layout analysis of printed documents has been produced in the end of the XX$^{th}$ century. Several algorithms have been proposed such as the X-Y Cut algorithm \citep{Nagy1992APD}, the Docstrum algorithm \citep{OGorman1993TheDS} or the Voronoi diagram based algorithm \citep{Kise1999OnTA}. Furthermore, the processing of handwritten documents requires specific techniques, such as the "droplet" technique to identify text line by  \citet{4378732}, or neural networks as in \citet{chen2017convolutional}, where each pixel is labelled as text or not.

Existing Logical Layout Analysis systems make use of various methods that go from heuristic systems to more recent architectures using neural networks. Some heuristic systems use grammars such as stochastic or attributed grammars, where the document is represented as a string of symbols, e.g. \citet{docstructure}. In their work, the grammar describes multiple production rules, each associated with a logical label. The string of symbols is then parsed by the grammar in order to extract logical labels. Other systems, such as LA-PDFText \citep{lapdftext} or DeLoS \citep{delos}, use rules that state the condition a physical block must meet to be given a logical label. For instance, DeLoS system uses first-order predicates in order to infer the logical category of a physical block.

While heuristic systems provide good results, they are often dedicated to specific layouts, and need to be adapted to work on other layouts. To tackle this problem, \citet{Klampfl2013AnUM} created a system for Logical Layout Analysis on scientific articles in PDF format that combines heuristic rules with unsupervised-learning models such as k-means or Hierarchical Agglomerative Clustering (HAC). This system is made up of several detectors, each learning geometrical and textual features from the document in order to identify a specific logical label. Some rules using text occurrences are also used to help the model, such as finding the keywords "Table" or "Fig." to identify table or figure blocks. 

More recent works use neural networks for Logical Layout Analysis. As noted by \citet{akl-etal-2019-fintoc}, Convolutional Neural Network (CNN) or Long Short-Term Memory (LSTM) architectures work better than classical neural networks because of the sequential nature of the documents. This task also benefits from the use of word-embeddings such as GloVe \citep{pennington-etal-2014-glove},  fastText \citep{bojanowski2016enriching} or Flair \citep{akbik2018coling} which give a better encoding of textual data than simple one-hot encoding, as in \citet{8946046}. Neural network systems can be trained on big data sets such as the Publaynet data set \citep{Zhong2019PubLayNetLD} or the Medical Articles Record Groundtruth (MARG) for physical and Logical Layout Analysis purposes.

Considering the task of processing historical documents, several small data sets exist such as the DIVA-HISDB data set \citep{Simistira2016DIVAHisDBAP} which contains 150 annotated pages of three different medieval manuscripts or the European Newspapers Project data set \citep{7333898} which contains 528 documents. Other data sets in non-European languages exist, such as the PHIBD data set \citep{phibd}, which contains images of 15 Persian historical and old manuscript, and the HJdata set \citep{Shen2020ALD}, which contains 2271 Japanese newspapers published in 1953, which was generated in a semi-automatic way. All of these data sets are too small to be used for Machine Learning or neural network approaches.

The works of \citet{crfarticle} deal with the task of article segmentation by a Conditional Random Field (CRF) model with heuristic rules to perform logical analysis. First the CRF model labels pixel as titles, text lines, or horizontal and vertical separators, then heuristics rules describing usual article layouts are applied to that classification. In both cases, bad results were caused by the quality of the scan or the quality of the OCR output. On the other hand, \citet{riedl-etal-2019-clustering} deal with article segmentation by looking at the similarity between segments of texts. These segments are computed either by using the Jaccard coefficient and their word distribution or by computing the cosine similarity between word-embeddings. The similarity between blocks is then computed using the TextTiling algorithm \citep{texttiling}.

Most common approaches to LLA are not suited for historical documents because the document layout changes over time. For example, the layout and structure of an advertisement in the same newspaper can display important changes over several years. Logical layout analysis systems applied to historical documents must then account for the diachronic aspect of their layouts and adapt to the changes. \citet{Barman2020CombiningVA} propose a system that goes beyond usual logical labels by labelling physical block as either Serial, Weather Forecast, Death Notice and Stock Exchange Table. To do so, their system combines visual and textual features using the word-embedding representation of each word and its coordinates on the page. Their results show that combining textual and visual features provide better results in most cases than using just one of them. Textual features are also more efficient to deal with the diachronic aspect of documents because they are more stable over time than visual features.

The rest of the article is organised as follows: Section \ref{methodology} presents our train and test data sets, the methodology that we propose and describe the three models compared in this article. Section \ref{evaluation} proposes an evaluation of the three models and compares their results. Finally, we propose our conclusion in Section \ref{conclusion}.

\section{Methods\\}
\label{methodology}

Our aim is to attribute logical layout labels to both TextBlock and TextLine tags in documents. 
For our experiment, first we design a rule-based system to perform the Logical Layout Analysis. This system is created using ad hoc rules and observations on the documents of the data set. Then we compare the results of this rule-based system with those of a Rule Learning model and a Machine Learning model.

In the following section we present our data set. Section \ref{label-tagset} presents the tagset that is used, and section \ref{feature-extraction} presents the features set that we have defined for the TextBlock and TextLine elements. Then, section \ref{pipeline} explains the general processing pipelines. Section \ref{rulebased} explains the construction of the rule-based system, while sections \ref{rulelearning} and \ref{machinelearning} present the Rule Learning and Machine Learning models that we used.

\subsection{Data set\\}
\label{data set}

\begin{figure}[htbp]
    \centering
    \caption{Excerpt of the first page of the second issue of the communist newspaper \textit{Le Semeur} published on the 23rd of April 1932}

    \includegraphics[width=0.86\textwidth]{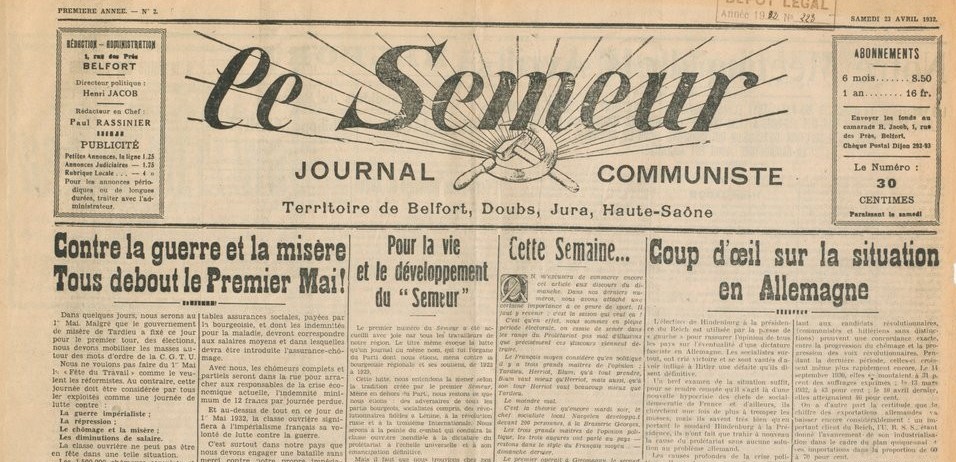}
    \label{fig:newspaper_page1}
\end{figure}

We have processed a data set of press and magazine documents published in the first half of the twentieth century from the "Fond régional: Franche-Comté" catalogue, available on Gallica, the digital archive of Bibliothèque Nationale de France (\url{https://gallica.bnf.fr}). The previous study by \citet{nicolas_gutehrle_2021_5752440} uses the same data set. Figure \ref{fig:newspaper_page1} shows an example of the first page of a newspaper with complex physical layout. It contains the header of the first page and several articles with titles and text content. From this catalogue, we selected documents that had an OCR quality measure greater than 90\%. This data set was then split into a train and a test data set. 

 The train and test data sets are designed to cover as much as possible the various physical layouts that exist in the "Fond régional: Franche-Comté" catalogue. We have divided them into three layout types: 


\begin{table}[H]
    \centering
    \begin{tabular}{cl}
         \textbf{1c :} &  documents where the text is displayed in one column, as in books;\\
         \textbf{2c :} &  documents where the text is displayed into two columns;\\
         \textbf{3c :} & documents where there are at least 3 columns of text, as in newspapers.
    \end{tabular}
\end{table}

As shown in Table \ref{tab:stattraintestset}, our train set contains 15 collections of documents, which amount to a total of 48 documents, whereas the test set contains 6 collections and a total of 6 documents. Table \ref{tab:distlayouttraintest} shows the distribution of documents in the train and test data sets across the three layout types.

\begin{table}[htbp]
\small
\centering
\caption{Description of the train and test data sets}

\begin{tabular}{lrrrrrr}
\hline
\textbf{Data set} & \textbf{Newspapers} & \textbf{Issues} & \textbf{Text blocks} & \textbf{Text lines} & \textbf{Words} & \textbf{Pages} \\ 
\hline
Train   & 15                 & 48           & 4~608             & 51~815           & 338~583      & 368         \\
Test    & 6                  & 6            & 1~445             & 8~836            & 63~343       & 52 \\
\hline
\end{tabular}
\normalsize
\label{tab:stattraintestset}
\normalsize
\end{table}

\begin{table}[htbp]
\small
\centering
\caption{Number of documents per layout in the train and test data sets}

\begin{tabular}{lrrrr}
\hline
\textbf{Data set / Layout} & \textbf{1c} & \textbf{2c} & \textbf{3c+} & \textbf{Total} \\ \hline
Train   & 18 & 5  & 25 & 48  \\
Test    & 2  & 2  & 2 & 6 \\ \hline
\end{tabular}

\label{tab:distlayouttraintest}
\normalsize
\end{table}

The documents in the corpus cover three general topics: Catholicism, Resistance and News. The documents of the Catholic topic were published between 1900 and 1918. Most of them, such as \textit{Bulletin paroissial de Censeau} or \textit{Petit Écho de Sainte-Madeleine}, are bulletins of small parishes. As such, they focus mainly on the local religious life, although they sometimes discuss national and international events such as WWI. The documents from the Resistance topic, such as \textit{La Haute-Saône libre} or \textit{La Franc-Comtoise}, were published between 1939 and 1945 by Resistance fighters. As such, their main goal is to relay information about the ongoing local and international events of WWII. Finally, the documents of the News topic were published in the 1930s and focus on local and national events. Some are apolitical such as \textit{Le Franc-Comtois de Paris}, while others have a political label. For instance, \textit{Le Semeur} and \textit{Le Front Comtois} are left-wing newspapers whereas \textit{Vers l’Avenir} is a right-wing Catholic newspaper. 

The French language used in these documents is not very different from modern French. However, we notice some variations in the written styles between the three topics. The written style in the Catholic document is formal and literary and uses many religious metaphors. On the other hand, the written style in the News document is mostly standard, although sometimes formal. Sentences are shorter and use simpler tenses than the Catholic documents. This simplification of the writing style is even more prominent in Resistance documents. The difference in the writing style between documents can first be explained by their domain: religious text should be more literary than newspapers or Resistance periodicals. This difference can also be explained by the size of the documents. Catholic documents are the longest in the corpus, with more than 10 pages on average. As such, their text can be more elaborate. On the other hand, News and Resistance documents are respectively four and two pages long on average. Their text is factual and concise in order to convey a lot of information in the limited space they have.

All the documents are stored in the XML ALTO format, which contains descriptions of their physical layout and the text content obtained from OCR. As such, the files already provide the physical layout analysis and the reading order of the documents.
The XML ALTO format provides the text content and physical layout of documents in the following manner: the OCR output for the whole document is available in a PrintSpace tag. Lines of text are contained in TextLine tags, which in their turn contain String tags for words and SP tags for spaces. TextLine tags are grouped into blocks in TextBlock tags. Sometimes, TextBlock tags are also grouped into ComposedBlock tags. TextBlock and TextLine tags have the following attributes:

    

\begin{table}[H]
    \centering
    \begin{tabularx}{\textwidth}{lX}
         \textbf{Id : } & the tag's identifier; \\
         \textbf{Height, Width :} &  the text height and width; \\
         \textbf{Vpos :} & the vertical position of the text on the page. The higher the value, the lower the word is on the page; \\
         \textbf{Hpos :} & the horizontal position of the text on the page. The higher the value, the further on the right the text is on the page; \\
         \textbf{Language :} &  the language of the text (only for TextBlock tags).\\
    \end{tabularx}

\end{table}

Besides the attributes listed above, a small portion of the TextBlock tags also have a Type attribute. This attribute is useful, when present, as it contains the logical labels of the lines in the block. It appears most often for tables or advertisements. However, the Type attribute is rarely present in our data set. As shown on Table \ref{tab:statblocktypetraintestset}, nearly 98 \% of the TextBlock tags in the train and the test data sets do not have a Type attribute.

\begin{table}[htbp]
\small
\centering
\caption{Type attribute distribution on TextBlock tags in the train and test data sets}

\begin{tabular}{l|cc|cc}
\hline
               & \multicolumn{2}{c|}{\textbf{Train}} & \multicolumn{2}{c}{\textbf{Test}} \\
\textbf{Type attribute} & \textbf{Count}     & \textbf{Percentage}    & \textbf{Count}    & \textbf{Percentage}    \\ \hline
No attribute   & 4~514      & 97.96        & 1~423     & 98.48        \\
illegible      & 79        & 1.71         & 15       & 1.04         \\
titre1         & 15        & 0.33         & 0        & 0             \\
advertisement  & 0         & 0             & 4        & 0.28         \\
table          & 0         & 0             & 2        & 0.14         \\
textStamped    & 0         & 0             & 1        & 0.07         \\ \hline
\end{tabular}
\label{tab:statblocktypetraintestset}
\normalsize
\end{table}


\subsection{Logical Layout Tagset\\}
\label{label-tagset}

To perform the Logical Layout Analysis of the documents, we define the following annotation tagset:

\begin{center}
    \begin{tabular}{cl}
         \textbf{TextBlock labels : } &  Text, Title, Header, Other \\
         \textbf{TextLine labels : } & Text, Firstline, Title, Header, Other
    \end{tabular}
\end{center}

The label "Firstline" must be understood as "first line of the paragraph". Thus, any TextLine tag labelled Firstline will indicate the beginning of a paragraph. A small portion of the TextBlock and TextLine tags correspond to elements that are not relevant for our study, such as images, tables or advertisement. Those elements were labelled as "Other" and are ignored for the evaluation.

The whole data set was manually annotated by a single annotator, then split into a train and a test data set. Table \ref{tab:lineclasseswholecorpus} shows the label distribution in the data sets. The train set was used to produce the rule set of our rule-based system, and also to train the Rule-learning and Machine-learning algorithms. The test set was used for the final evaluation of the system.

\begin{table}[htbp]
\small
\centering
\caption{TextBlock and TextLine tags label distribution over the train and test data sets}

\begin{tabular}{ll|cc|cc}
\hline
&          & \multicolumn{2}{c|}{\textbf{Train}} & \multicolumn{2}{c}{\textbf{Test}} \\
& \textbf{Label}   & \textbf{Count}   & \textbf{Percentage} & \textbf{Count}  & \textbf{Percentage} \\ 
\hline

\multirow{4}{*}{\rotatebox{90}{\textbf{TextBlock}}} & Text &      2~064 & 45.724 & 1~102 &  80.203 \\
& Title &       429 & 9.503 & 90 &   6.550 \\
& Header &       333 & 7.377 & 53 &   3.857 \\
& Other &       1~686 & 37.35 & 128 &   9.314 \\
\hline

\multirow{5}{*}{\rotatebox{90}{\textbf{TextLine}}} & Text      & 36~272  & 70.138     & 6~648  & 75.881     \\
& Firstline & 9~785   & 18.921     & 1~563  & 17.840     \\
& Title     & 1~820   & 3.519      & 234    & 2.670      \\
& Header    & 740     & 1.430      & 115    & 1.312      \\
& Other     & 3~098     & 5.989     & 201   & 2.293     \\
\hline

\end{tabular}
\label{tab:lineclasseswholecorpus}
\normalsize

\end{table}


\subsection{TextBlock, TextLine and Document features\\}
\label{feature-extraction}

The first step of our processing pipelines consists in extracting and calculating sets of features from XML ALTO documents. These features describe three levels: TextLine, TextBlock and Document level. Table \ref{tab:tabfeatures} presents all the features with their descriptions and levels. The information on these features for all document elements, in the form of matrices, is used as input for the three models presented in Sections \ref{rulebased}, \ref{rulelearning}, \ref{machinelearning}.

\begin{table}[H]
\small
\caption{Features used by the algorithm. The features marked with an * are used exclusively by the rule-based algorithm}

\begin{tabularx}{\textwidth}{|lp{4.7cm}X|l|l|l|}
    \hline
      & \textbf{Feature} & \textbf{Description} & \rotatebox{90}{\textbf{\footnotesize TextLine}} &  \rotatebox{90}{\footnotesize \textbf{TextBlock }} & \rotatebox{90}{\footnotesize \textbf{Document }} \\ 
      \hline
     1 & \textit{page} & page number of the page containing the element & X & X & \\
     \hline
     2 & \textit{blockType} & type of the block
     & X & X &\\
     \hline
     3 & \textit{wordCount} & number of words & X & X & \\
     \hline
     4 & \textit{precedingSpace, followingSpace} & spaces above and below the element & X & X & \\
     \hline
     5 & \textit{height, width} & height and width values of the line & X &  &  \\
    \hline
     6 & \textit{hpos, vpos} & coordinates of the line on the page, i.e. its horizontal and vertical position & X &  & \\
     \hline
     7 & \textit{diffHpos} & difference between \textit{hpos} and the median \textit{hpos} value in the block & X &  &\\
     \hline
     8 & \textit{firsthpos, firstvpos} & coordinates of the first line of the block  &  & X & \\
     \hline
     9 & \textit{lasthpos, lastvpos} & coordinates of the last line of the block  &  & X &\\\hline
     10 & \textit{linecount} & number of lines  &  & X & \\
     \hline
     11 & \textit{wordRatio} & number of words by line  &  & X & \\ 
     \hline
     12* & \textit{medHeight, medWidth} & median line height and line width  &  & X & X \\
     \hline
     13* & \textit{medHpos, medVpos} & median \textit{hpos} and \textit{vpos} values in the block  &  & X & \\
     \hline
     14* & \textit{medWordCount, medLineSpace} & median number of words by line and the median space between lines in the block  &  & X & X \\
     \hline
     15* & \textit{medBlockHeight, medBlockWidth} & median line height and block height and width &  &  & X\\
     \hline
     16* & \textit{medBlockSpace} & median space value between blocks &  &  & X\\
     \hline
     17* & \textit{thirdQuartileLineSpace} & third quartile of line space values in the document &  &  & X\\
     \hline
     18* & \textit{medWordRatio, medLineCount} & median number of words by line and median number of line by block in the document &  &  & X\\ \hline

     19 & \textit{capitalProp, digitProp} & proportion of capital letters and digits & X & X & \\
     & \textit{nonAlphaProp} & proportion of non-alphanmeric characters &  &  & \\    
    \hline
     20 & \textit{stwCapital, stwDigit} & True if the line starts either by a capital letter or a number, False otherwise & X &  &\\
    \hline
    21 & \textit{endsPunct} & True if the line ends with a punctuation, False otherwise & X &  &\\
    \hline

     22 & \textit{headerMark1} & True if the element contains the word "Page" or a dash sign. False otherwise.& X & X & 
    \\\hline
    23 & \textit{headerMark2} & True if the element contains a date, a currency, an address. False otherwise.& X & X &    \\
    \hline
     24 & \textit{simTitle} & similarity of the line with the title of the document, calculated by the Levenshtein distance & X &  &\\
     \hline
     25 & \textit{simHeaderSet} & highest similarity of the line with the words contained in the header words set, calculated by the Levenshtein distance & X &  &\\
     \hline
     
\end{tabularx}
\normalsize
\label{tab:tabfeatures}
\end{table}

The header words set is used for the calculation of the \textit{simHeaderSet} feature. It has been created by observing the different types of headers in the data sets and consists of the following words or phrases: \textit{Rubrique Locale, Gérant, Publicité, Abonnement, Envoyez les fonds, Conservez chaque numéro, Rédacteur, Directeur, Numéro, Chèque postal, Dépôt, Achat-Vente-Echange, Annonce, Imprimerie, En vente partout, Paraissant}. \\


\subsection{Processing pipelines\\}
\label{pipeline}

In this section we present the two pipelines that we propose for the processing of XML ALTO documents for Logical Layout analysis. First, we present the rule-based system, and then we explain how it can integrate the Rule Learning or Machine Learning models.

The first step of the processing pipeline consists in the extraction of features from the XML ALTO document at the TextLine, TextBlock and Document levels, as described in Section \ref{feature-extraction}.  
We store these features into two matrices for TextLine and TextBlock features and in a dictionary for Document features. Each row in the matrices represents either a TextLine or a TextBlock tag and each column is a corresponding feature.

\subsubsection{Rule-based system\\}

For the rule-based system, the second step attributes logical labels to TextBlock tags. Labelling TextBlock before TextLine is important because the presence of a Type attribute in TextBlocks can help label the lines inside these blocks. The goal of this step is to add a Type attribute to every TextBlock. To do so, we process the TextBlock feature matrix from the previous step by applying sets of annotation rules, one for each possible logical label. A TextBlock is only processed if it doesn't already have a Type attribute. Because the sets of rules are applied independently from each other, a same TextBlock can obtain multiple labels. Another set of rules is then applied to solve such conflicts and keep only one possible logical label for each TextBlock, which is then set as the value of the Block's Type attribute in the feature matrix. 

The third step attributes logical labels to TextLine tags. Every TextLine is by default labelled as Text. The system then applies rules to identify the other labels. First, any TextLine in a Title or a Header block inherits the same label. Then, any TextLine contained in a TextBlock is processed by a set of rules in order to identify Firstlines and possible missing Titles. Similarly to the previous step, rules are applied independently from each other, resulting sometimes in conflicting predictions. The TextLine feature matrix is processed a second time to solve conflicting predictions and keep only one possible label for each TextLine tags. This step also controls that any line that follows a Title is labelled as Firstline and that the first line of the document is labelled as Title if it not already labelled as Header. 

\begin{figure}[htbp]
    \centering
    \caption{Processing pipeline of the rule-based system}
    \includegraphics[width=.9\textwidth]{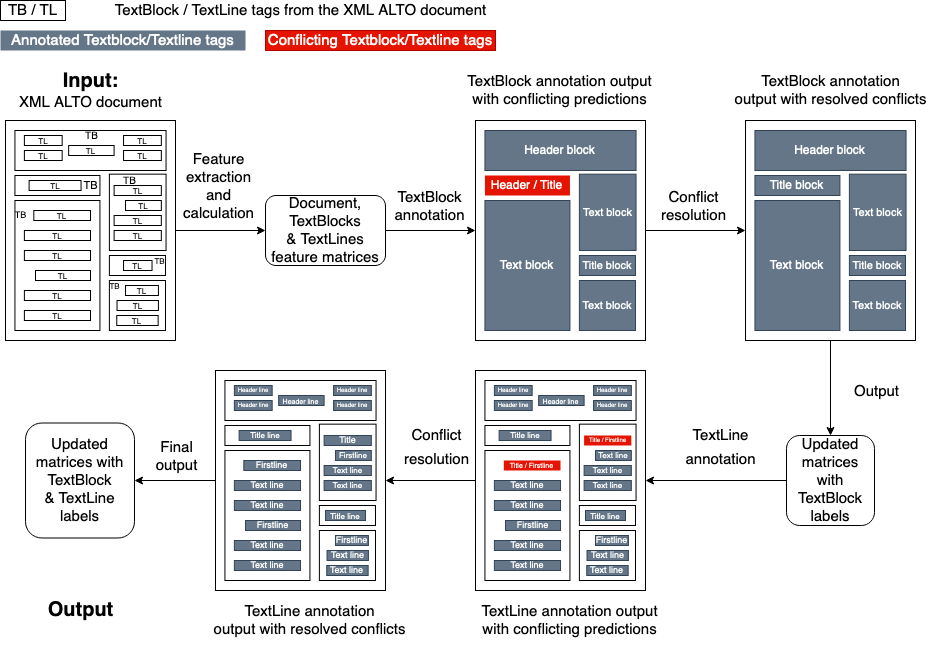}
    \label{fig:algorithm_structure}
\end{figure}

The rule set of our rule-based system uses all features presented in Table \ref{tab:tabfeatures}. 
The algorithm finally outputs the three feature matrices, where the TextBlock and TextLine matrices are updated with the annotations of steps 2 and 3. Figure \ref{fig:algorithm_structure} shows the main steps of this pipeline.

\subsubsection{Rule Learning and Machine Learning pipelines\\}

\begin{figure}[htbp]
    \centering
    \caption{Processing pipeline of the Rule Learning and Machine Learning algorithms}

    \includegraphics[width=0.7\textwidth]{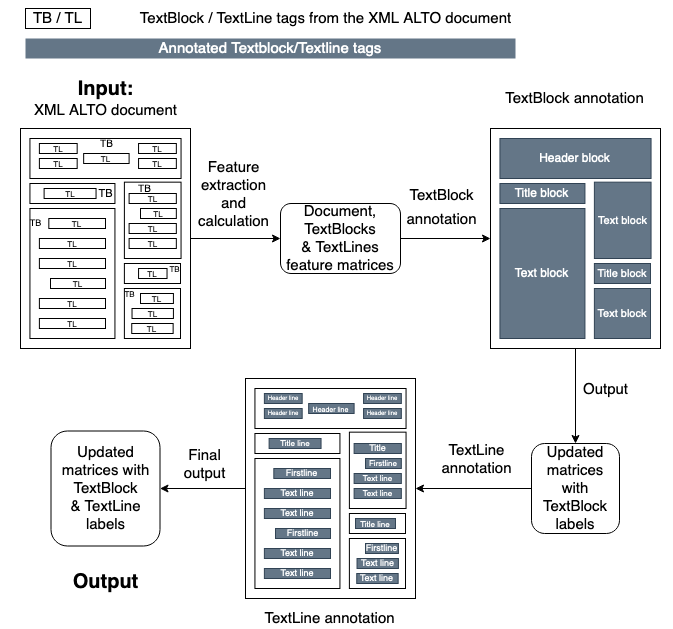}
    \label{fig:algorithm_structure2}
\end{figure}

The processing pipeline which integrates the Rule Learning and Machine Learning methods differs from the rule-based one as follows. The second step attributes logical labels to TextBlock tags and updates the feature matrices accordingly. The last step attributes logical labels to TextLine tags and outputs the final updated matrices. The steps 2 and 3 rely on Rule Learning or Machine Learning models. Unlike our rule-based system, however, there is no need for conflict resolution steps, as the Rule Learning and Machine Learning algorithms output the final prediction. Figure \ref{fig:algorithm_structure2} shows the main steps of this pipeline.

To train the Rule Learning and Machine Learning algorithms, we use the same set of features, but we exclude features 12 to 18. These features contain values that are calculated using the data of the other features. They are used only by the rule-based system, and not by the other algorithms as they do not provide any new data.

\subsection{Rule-based system\\}
\label{rulebased}

To construct the rule-based system, we define sets of rules for the annotation of TextBlock and TextLine tags. To design the rules, we used heuristics and observations that we made on the train data set. For instance, we observed that the biggest titles in the documents start with a capital letter and are surrounded by important spaces. Then, we translated these patterns into rules in the form of conditions on the values of the features that must be that must be verified to trigger the annotation of the element with the corresponding label.
The obtained set of rules is applied to documents regardless of the layout category they belong to.

Identifying Text and Title blocks relies on geometric and morphological features, whereas identifying Header blocks relies on semantic features.
While designing the rules for the annotation of TextBlock elements we have taken into consideration the following observations:

\begin{itemize}
    \item Text blocks contain relatively more lines and more words than other blocks (titles or headers) in the document.
    \item Title blocks are TextBlock tags that contain few lines, usually not more than 3. The role of a title is to introduce the topic of a text section, thus a Title block should be surrounded by Text blocks. The space around that block should also be important, in order to stand out with the surrounding blocks.
    \item A Text block should have a smaller height than a Title block. As such, if there is a confusion between Text and Title block, we use the height of the block to distinguish between the two. 
    \item Headers contain very specific information about the document, such as its title, its price, a date or the publisher's name. This information is displayed with keywords and sentences that are recurrent across multiple pages and documents. 
    \item Header blocks are only located at the top of a page, generally in the first four lines of a page. Small blocks at the top of a page are most likely to be Headers.
    \item On the first page of a document, the header can be much longer because it contains more information. We consider that it can be up to 30 lines.
\end{itemize}

Naturally, TextLine tags that are contained in a Title or Header block inherit this annotation. TextLine tags that appear between two Header lines are also annotated as Header. Furthermore, to find Firstline and Title lines that have not yet been identified, we apply sets of rules that rely on geometric and morphological features, taking into account the following observations:

\begin{itemize}
    \item The first line of a page or immediately after a Title should be labelled Firstline, if it starts with a capital letter.
    \item The first line of a paragraph always starts with a capital letter, and is often indented, i.e. with Hpos value greater than the other TextLines in the block. 
    \item Firstlines that are not indented can be identified if the line that precedes them is shorter, indicating the end of the previous paragraph.
    \item Title lines are surrounded by relatively more space in order to stand out from other text sections. The smaller the title is, the less important the space around it is.
    \item  Small titles (e.g. sections in an article) usually contain more capital letters then the rest of the text and are center-aligned. 
\end{itemize}

Tables \ref{algo-block} and  \ref{algo-lines} present all annotation rules for TextBlock and TextLine tags and their corresponding annotation labels, where $B$ is a TextBlock in a document $D$ and $L$ is a TextLine in $B$. The last two rules in each table solve conflicting annotations.

\begin{table}[H]
\small
\caption{TextBlock annotation rules and conflict resolution rules of the rule-based system}
\begin{tabularx}{\textwidth}{lXl} \hline
     \textbf{Rule} & \textbf{Condition} & \textbf{Label} \\ \hline
     1 & \textit{(B.linecount > D.medLineCount)} or \textit{(B.wordCount > D.medWordCount/3)} & Text \\
     2 & Previous and next TextBlocks are Text and \textit{(B.linecount < D.medLineCount)}
  and \textit{(B.medHeight < D.medBlockHeight)} & Text \\ \hline

     3 & Previous and next TextBlocks are Text and $B$ is not Text and \textit{(B.linecount < 4)} and \textit{(B.precedingSpace > D.medBlockSpace)
    or (B.followingSpace > D.medBlockSpace)}  & Title \\ \hline
    4 & \textit{B.page = 1} and for any of the first 30 lines of $B$: \textit{simHeaderSet > 0.9} or \textit{simTitle > 0.9} or \textit{headerMark1} or \textit{headerMark2} or \textit{ctnTotal} & Header \\
    5 & \textit{B.page > 1} and for any of the first 4 lines of $B$: \textit{simHeaderSet > 0.9} or \textit{simTitle > 0.9} or \textit{headerMark1}  & Header \\ \hline
    
    6 & Conflicting annotation: Header and (Text or Title):  \\
     & \textit{(B.linecount < 15)} and \textit{(B.wordCount < 50)} & Header \\ 
     & Otherwise & Text / Title \\
     \hline
     7 & Conflicting annotation: Text and Title: \\
     & \textit{B.medHeight > D.medBlockHeight / 2} & Title \\
     & Otherwise & Text \\
     \hline
\end{tabularx}
\normalsize
\label{algo-block}
\end{table}

\begin{table}[H]
\small
\caption{TextLine annotation rules and conflict resolution rules of the rule-based system}
\begin{tabularx}{\textwidth}{lXl} \hline
     \textbf{Rule} & \textbf{Condition} & \textbf{Label} \\ \hline

     1 & L.\textit{precedingSpace} = 0 and L.\textit{followingSpace} > D.\textit{medLineSpace} and L.\textit{simTitle} < 60 
     and L.\textit{simHeaderSet} < 60 and L.\textit{stwCapital} & Title \\
     2 & L.\textit{wordCount} < B.\textit{medWordCount} and L.\textit{precedingSpace} > D.\textit{thirdQuartileLineSpace} and L.\textit{followingSpace} > D.\textit{thirdQuartileLineSpace} & Title \\
     3 & L.\textit{capitalProp} > 10 and L.\textit{wordCount} < B.\textit{medWordCount} and L.\textit{height} < B.\textit{medHeight} and (L.\textit{precedingSpace} > D.\textit{thirdQuartileLineSpace} or L.\textit{followingSpace} > D.\textit{thirdQuartileLineSpace}) & Title \\
     4 & L.\textit{diffHpos} > 104 and L.\textit{capitalProp} > 0 and L.\textit{precedingSpace} > D.\textit{medLineSpace} and L.\textit{followingSpace} > D.\textit{medLineSpace} & Title \\
     \hline
     5 & L.\textit{hpos} > B.\textit{medHpos} and L.\textit{diffHpos} < 105 and (L.\textit{stwCapital} or L.\textit{stwDigit}) & Firstline \\
     6 & L.\textit{width} < B.\textit{medWidth} and L.\textit{wordCount} < B.\textit{medWordCount} and L.\textit{hpos} < B.\textit{medHpos} & Lastline \\
     7 & Previous TextLine is LastLine and L.\textit{stwCapital} and L.\textit{followingSpace} < B.\textit{medLineSpace} & Firstline \\
     8 & Previous TextLine is not Lastline and L.\textit{stwCapital} and L.\textit{precedingSpace} > B.\textit{medLineSpace} and L.\textit{followingSpace} < B.\textit{medLineSpace} & Firstline \\
     9 & Previous TextLine is not Lastline and L.\textit{stwCapital} and L.\textit{hpos} > B.\textit{medHpos} & Firstline \\
    \hline
    10 & None of the rules 1-9 above is True & Text \\
    \hline
    
     11 & Conflicting annotation: Header and other label:  \\
     & Previous TextLine is Header and next TextLine is Header & Header \\
     \hline
     12 & Conflicting annotation: Title and FirstLine: \\
     & L.\textit{followingSpace} < B.\textit{medLineSpace} and L.\textit{capitalProp} < 15 & Title \\
     & Otherwise & Firstline \\
     \hline
\end{tabularx}
\label{algo-lines}
\normalsize
\end{table}


\subsection{Rule Learning\\}
\label{rulelearning}

Rule Learning algorithms are Machine Learning algorithms that aim at identifying rules from a data set. In a classification task, a rule-learning model induces from the data set the rules that allow to classify a particular sample. For our experiment, we have selected the RIPPER algorithm for two reasons: firstly, it is considered to be the state of the art of rule-induction systems \citep{sammut-encyclopedia}, and secondly, for its ability to produce human-readable rule sets.

\subsubsection{Description of the RIPPER algorithm\\}

RIPPER (Repeated Incremental Pruning to Produce Error Reduction) \citep{cohen1995repeated} is a rule induction algorithm which improves upon the Incremental Reduced Error Pruning algorithm (IREP) \citep{furnkranz1994incremental}. RIPPER discovers rules in a sequential covering manner: for a positive class $P$, it identifies the rules that best cover the examples of $P$ in the data set. Every element in the data set covered by the rule, be it positive or negative, is then removed. These operations are repeated until the rule set cannot grow any more or another condition is met.

In order to construct the rule set, RIPPER first randomly divides the training set into a \textit{growing} and a \textit{pruning} set. The growing set is used to identify rules and corresponds roughly to 2/3 of the training set. The pruning set is used to test the rules and prune them. Building a rule consists of the two following steps: \textit{Growing} and \textit{Pruning}.

Rules in RIPPER are sets of conditions combined using the AND operator. The growing step of a rule starts with an empty rule. The algorithm loops over every possible condition in the data set, i.e. every possible value for each given feature. It then adds the condition which provides the highest
information gain to the rule. The information gain is calculated with the same metric used by the 
First Order Inductive Learner (FOIL) algorithm \citep{Quinlan2005LearningLD}. The algorithm keeps adding conditions to the rule until it only covers examples of the positive class. Once a rule has stopped growing, its quality is evaluated with the following metric: \[ruleQuality = (P-N) / (P+N)\] where $P$ and $N$ are respectively the number of positive and negative examples covered by the pruned rule in the pruning set. 

The pruning step is then initiated. It consists in removing one by one every condition in the rule, from the newest to the oldest, while evaluating its quality with the same metric. The version of the rule having the best quality is kept.

When a rule is grown and pruned, RIPPER computes the description length (DL) of the new rule. The DL indicates the complexity of a rule and corresponds to the sum of the number of bits required to encode the rule and the examples of the positive class the rule fails to cover. Finally, every example in the growing set covered by the newly built rule is removed. These steps are repeated until one of the following conditions is met:

\begin{itemize}
    \item the rule set covers every instance of the positive class; 
    \item the error rate is above 50 \% since the addition of the last rule;
    \item the rule has reached a specified complexity threshold.
\end{itemize}

The complexity threshold is reached when the DL of the last added rule is $d$ bits bigger than the smallest DL in the rule set. By default, $d$ is 64 bits. After creating the rule set, RIPPER iterates from the newest to the oldest rule in the rule set and checks if the rule can be removed without increasing the total description length. 

Finally, the optimisation step is initiated. It consists in creating two new versions of each rule: a replacement rule and a revision rule. The replacement is a completely new rule whereas the revision rule is created by adding new conditions to the existing rule. The algorithm then selects the version of the rule with the smallest description length. The optimisation step may be run $k$ times, where $k=2$ by default.

For our experiment, we used the python library \textit{wittgenstein} which implements the RIPPER algorithm. The main hyperparameters of this implementation are:

\begin{center}
\begin{tabularx}{\textwidth}{lX}
     \textbf{prune\_size :} &  the size of the prune set. The default value is .33 \\
     
     \textbf{k :} & the number of optimisations to run. The default value is 2 \\
     \textbf{dl\_allowance :} & the allowed size for description length. The default value is 64 \\
     
    \textbf{n\_discretize\_bins :} & specific to the \textit{wittgenstein} implementation. It automatically detects and discretises continuous features in the training set. This hyperparameter controls the size of each bin. The default value is 10.
\end{tabularx}
\end{center}

As the \textit{wittgenstein}'s implementation of RIPPER can only deal with binary classification, we trained one model for each label in our data set for both the TextBlock and TextLine annotation tasks. For each RIPPER model, we performed a GridSearch with the following combination to find the best hyperparameter values:

\begin{center}
\begin{tabular}{ll}
     \textbf{prune\_size :} &  .25, .33, .50 \\
     \textbf{k :} & 1, 2 \\
     \textbf{dl\_allowance :} & 32, 64, 128 \\
    \textbf{n\_discretize\_bins :} & 5, 10, 20, 30
\end{tabular}
\end{center}

Table \ref{tab:ripperhyperparameters} shows the best hyperparameter combinations for each RIPPER model we trained for the TextBlock and TextLine annotation tasks. 

For each logical label in the TextBlock or TextLine annotation task, we use the corresponding RIPPER model to predict the probability that the TextBlock or TextLine tag belongs to that logical label. The most probable label is then assigned to the tag. 

\begin{table}[htbp]
    \small 
    \centering
        \caption{Best hyperparameter combinations for the RIPPER algorithm on each class for TextBlock and Textline annotation tasks}

    \begin{tabular}{ll|cccc} \hline
     &   \textbf{Hyperparameter}      & \textbf{Header} & \textbf{Title} & \textbf{Firstline} & \textbf{Text} \\ \hline
     \multirow{4}{*}{\rotatebox{90}{\textbf{TextBlock}}} &   prune\_size         & .5     & .33   & & .25  \\
      &  k                   & 2      & 1     & & 2    \\
       & dl\_allowance       & 64     & 64    & & 64   \\
    &    n\_discretize\_bins & 10     & 10    & & 10  \\
    \hline
    \multirow{4}{*}{\rotatebox{90}{\textbf{TextLine}}} &   prune\_size        & .33    & .25   & .33                           & .5   \\
     &   k                & 1      & 1     & 1                             & 2    \\
     &   dl\_allowance      & 64     & 64    & 64                            & 64   \\
     &   n\_discretize\_bins & 10     & 10    & 10                            & 10   \\ \hline
    \end{tabular}
    \label{tab:ripperhyperparameters}
\end{table}


\subsection{Machine Learning\\}
\label{machinelearning}

Machine Learning represents the family of algorithms such as Support Vector Machine or Gradient Boosting which are able to learn patterns from the data on their own. Machine Learning algorithms are often quicker to train than rule-based systems. However, they are also harder to interpret as they do not output rules, nor explain their reasoning to classify a specific sample.

\begin{figure}[htbp]
    \centering
    \caption{Comparison of initial results of Machine Learning algorithms for TextBlock (left) and TextLine (right) annotations}
    
\subfigure{\label{fig:a}\includegraphics[width=75mm]{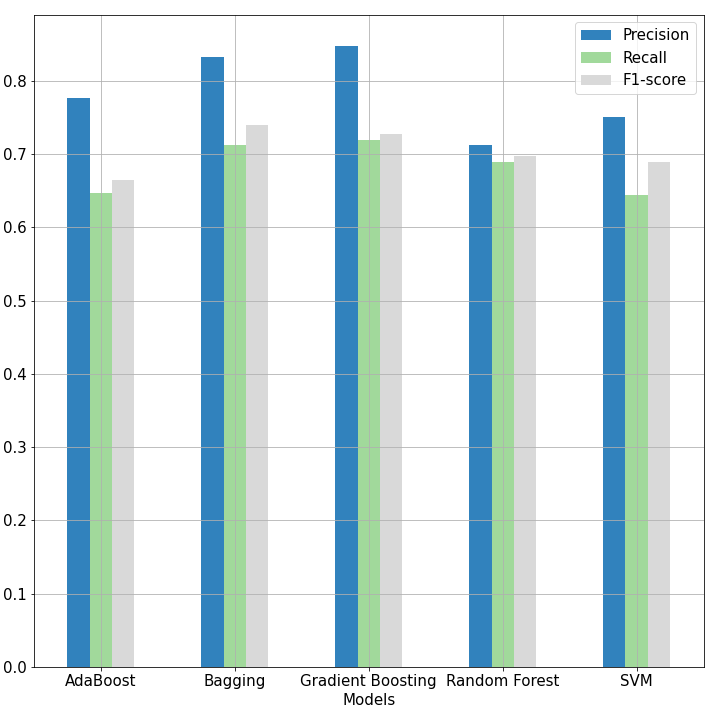}}
\subfigure{\label{fig:b}\includegraphics[width=75mm]{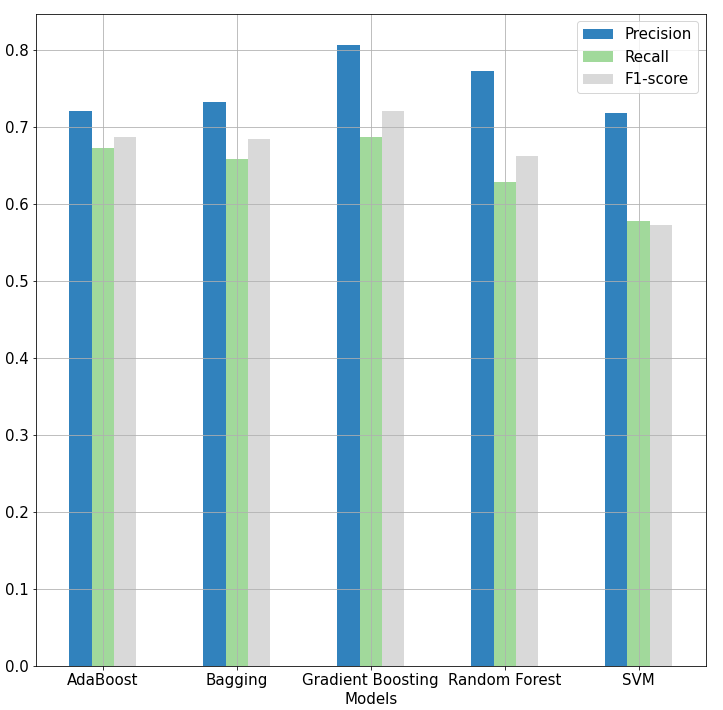}}

    \label{fig:graphmlresultscomparaison}
\end{figure}

In order to select the algorithm which is best suited for TextBlock and TextLine classification, we compared the performances of the following Machine Learning algorithms: \textit{Support Vector Machine (SVM)}, \textit{Bagging}, \textit{Random Forest}, \textit{AdaBoost} and \textit{Gradient Boosting}. We used the \textit{scikit-learn} implementation of these algorithms \citep{scikit-learn}, with their default hyperparameters on the same feature set that was used to train the RIPPER algorithm. Table \ref{tab:mlresultscomparaison} and Figure \ref{fig:graphmlresultscomparaison} compare the initial results obtained by these algorithms. The best result in each column is shown in bold. 

\begin{table}[htbp]
    \centering
        \caption{Results of Machine Learning models for TextBlock and TextLine annotation}

    \begin{tabular}{l|ccc|ccc} \hline
                 & \multicolumn{3}{c|}{\textbf{TextBlock}} & \multicolumn{3}{c}{\textbf{TextLine}} \\ 
    \textbf{Algorithm}       & \textbf{Precision}   & \textbf{Recall}  & \textbf{F1}     & \textbf{Precision}  & \textbf{Recall}  & \textbf{F1}    \\ \hline
    SVM          & 0.750       & 0.644   & 0.688  & 0.717      & 0.578   & 0.572 \\
    Bagging      & 0.833       & 0.712   & \textbf{0.740}  & 0.732      & 0.657   & 0.683 \\
    RandomForest & 0.712       & 0.689   & 0.697  & 0.773      & 0.628   & 0.662 \\
    AdaBoost     & 0.775       & 0.647   & 0.664  & 0.720      & 0.672   & 0.687 \\
    Gradient Boosting & \textbf{0.847} & \textbf{0.719} & 0.727 & \textbf{0.806} & \textbf{0.687} & \textbf{0.720} \\ \hline
    \end{tabular}
    \label{tab:mlresultscomparaison}
\end{table}

This initial comparison shows that Gradient Boosting is the best algorithm for both TextBlock and TextLine annotation in every evaluation except one. From these results, we trained two Gradient Boosting classifiers, one for TextBlock annotation and another for TextLine annotation. We performed a GridSearch for each task with the following sets of hyperparameters in order to find the optimal combination:

\begin{center}
    \begin{tabular}{ll}
        \textbf{n\_estimators :} &  100, 150, 200\\
        \textbf{learning\_rate :} & 0.01, 0.005, 0.001\\
        \textbf{max\_leaf\_nodes :} & 2,3,4,5,6,7\\
        \textbf{min\_samples\_split :} & 10, 20, 40, 60, 100\\
        \textbf{min\_samples\_leaf :} & 1,3,5,7,9\\
        \textbf{max\_features :} & 2,3,4,5,6,7\\
        \textbf{subsample :} & 0.7,0.75,0.8,0.85,0.9,0.95,1\\
    \end{tabular}
\end{center}

Table \ref{tab:gbhyperparameters} shows the best hyperparameter combinations for the Gradient Boosting algorithm identified by GridSearch for both tasks.

\begin{table}[htbp]
    \centering
        \caption{Best hyperparameter combinations for Gradient Boosting for TextBlock and TextLine annotation}

    \begin{tabular}{l|cc} \hline
    \textbf{Hyperparameter}      & \textbf{TextBlock} & \textbf{TextLine} \\ \hline
    n\_estimators       & 200       & 100      \\
    learning\_rate      & 0.005     & 0.01     \\
    max\_leaf\_nodes    & 7         & 7        \\
    min\_samples\_split & 40        & 100      \\
    min\_samples\_leaf  & 5         & 7        \\
    max\_features       & 6         & 7        \\
    subsample           & 0.85      & 1       \\ \hline
    \end{tabular}
    \label{tab:gbhyperparameters}
\end{table}


\section{Results and Discussion\\}
\label{evaluation}

To evaluate our rule-based system, as well as the Rule Learning and Machine Learning algorithms, we have run the three models on the test data set. In this section we present the results in terms of Precision, Recall and F1-score. Considering the Rule learning algorithm, we also present the set of rules that were obtained by the algorithm and compare this set with the one that we have defined in our rule-based system. Finally, we compare the results of all systems.

\subsection{Evaluation of the rule-based system\\}

Table \ref{tab:eval1} shows the Precision, Recall and F1 scores for the TextBlock and TextLine classification of the rule-based system.

\begin{table}[htbp]
\centering
\scriptsize
\caption{Precision, Recall and F1 score for TextBlock and TextLine annotation with the rule-based system}

\begin{tabular}{ll|ccc|ccc|ccc|ccc}
\hline
   &  & \multicolumn{3}{c|}{\textbf{Text}} & \multicolumn{3}{c|}{\textbf{Title}} & \multicolumn{3}{c|}{\textbf{Firstline}} & \multicolumn{3}{c}{\textbf{Header}} \\
& \textbf{Cat}  & \textbf{P}       & \textbf{R}      & \textbf{F1}     & \textbf{P}       & \textbf{R}      & \textbf{F1}     & \textbf{P}       & \textbf{R}      & \textbf{F1}       & \textbf{P}       & \textbf{R}      & \textbf{F1}      \\ 
\hline

 \multirow{4}{*}{\rotatebox{90}{\textbf{TextBlock}}} & 1c   & 0.947   & 0.938  & 0.942  & 0.312   & 0.357   & 0.333  & &&& 0.679   & 0.373   & 0.476   \\
& 2c   & 0.973   & 0.989  & 0.981  & 0.899   & 1.000   & 0.947  & &&& 1.000   & 0.271   & 0.411   \\
& 3c+  & 0.958   & 0.973  & 0.965  & 0.589   & 0.560   & 0.551  & &&& 0.500   & 0.250   & 0.333   \\ 
\cline{2-14}
& Mean & 0.959   & 0.966  & 0.962  & 0.600   & 0.639   & 0.610  & &&& 0.726   & 0.298   & 0.406   \\ 

\hline

 \multirow{4}{*}{\rotatebox{90}{\textbf{TextLine}}} & 1c   & 0.979   & 0.986  & 0.983  & 0.354   & 0.720   & 0.473  & 0.943    & 0.854    & 0.895    & 0.909   & 0.598   & 0.721   \\
& 2c   & 0.961   & 0.995  & 0.978  & 0.746   & 0.765   & 0.747  & 0.955    & 0.859    & 0.902    & 1.000   & 0.118   & 0.197   \\
& 3c+  & 0.975   & 0.992  & 0.983  & 0.703   & 0.702   & 0.702  & 0.952    & 0.877    & 0.913    & 0.500   & 0.400   & 0.444   \\ 
\cline{2-14}
& Mean & 0.969   & 0.991  & 0.979  & 0.595   & 0.733   & 0.639  & 0.949    & 0.861    & 0.902    & 0.803   & 0.348   & 0.435   \\
\hline
\end{tabular}
\normalsize
\label{tab:eval1}

\end{table}


TextBlock annotation rules perform best on documents from the 2c layout category. Title classification for TextBlocks performs with F1 score of 0.61 on average and 0.94 on documents from the 2c category. Header classification for TextBlocks provides a good Precision score (0.726) but with a low Recall (0.298).

Similarly to TextBlock annotation rules, TextLine annotation rules perform best on documents from the 2c category. Title identification performs worse on 1c documents, and obtains overall F1 score of 0.639 for all layouts. Firstline identification performs fairly well with an F1 score above 0.9. Header identification obtains a good Precision score (0.803) but with a Recall of 0.348. This means that header identification rules are insufficient and need to be completed to capture the various types of headers.

A first type of error comes from errors in the Block classification step. As any line in a Title or Header block inherits that annotation, the Precision of TextBlock annotation is an important factor for the overall performance of the algorithm. 

A second type of error is the confusion between Titles and First lines. Most Titles mislabelled as Firstline are short subsection titles. As such, they are similar to other text lines in terms of typography, and are hard to detect with the features we use. This confusion happens mainly in documents from the 2c and 3c+ categories. Other mislabelled Titles are one-line paragraphs such as greetings or signatures, or the beginning of a text section. Such lines have properties similar to Titles, being surrounded by important spaces and being either center or right-aligned. Extracting features about the font style of the line (bold, italics) and its alignment (left, center, right-aligned) could help solve this confusion.


\subsection{Comparison of RIPPER's annotation rules and the rule-based system\\}

In this section we present a comparison between the rules that are used in our rule-based system and the annotation rules that were learned by the RIPPER algorithm. We examine the sizes of the two rule sets and the overlap that exists between them.

\subsubsection{TextBlock annotation rules\\}

Table \ref{tab:ripperalgo-block} presents all annotation rules discovered by the RIPPER algorithm for TextBlock tags and their corresponding annotation labels, where $B$ is a TextBlock in a document $D$. Overall, RIPPER produced 26 rules, while in the rule-based system we produced 7, including the conflict resolution rules. 

\begin{table}[H]
\small
\caption{TextBlock annotation rules produced by the RIPPER algorithm}

\begin{tabularx}{\textwidth}{lXl} \hline
    \textbf{Rule} & \textbf{Condition} & \textbf{Label} \\ \hline
    1 & \textit{B.capitalProp => 55} and \textit{80 <= B.followingSpace <= 135}
     and \textit{135 <= B.width <= 368.25} & Title \\

    2 & \textit{B.height => 95} and \textit{801.25 <= B.width <= 985} and
     \textit{5 <= B.capitalProp <= 12.5} & Title \\
     
    3 & \textit{B.capitalProp => 55} and \textit{940 <= B.firstvpos <= 1545} and
     \textit{B.height <= 30} & Title \\
     
    4 & \textit{B.height => 95} and \textit{485 <= B.firstvpos <= 940} & Title \\

    5 & \textit{B.linecount <= 2} and \textit{B.height => 95} and \textit{940 <= B.firstvpos <= 1545}
    and \textit{79 <= B.wordRatio <= 96} & Title \\
    
    6 & \textit{B.capitalProp => 55} and \textit{B.wordRatio => 149} and \textit{B.digitProp <= 5}
    and \textit{B.lastvpos <= 900} & Title \\
    
    7 & \textit{B.capitalProp => 55} and \textit{80 <= B.followingSpace <= 135} and
    \textit{B.linecount <= 2} and \textit{4 <= B.wordCount <= 9} & Title \\
    
    8 & \textit{B.linecount <= 2} and \textit{B.digitProp <= 5} and \textit{B.height => 95}
    and \textit{175 <= B.followingSpace <= 225} & Title \\
    
    9 & \textit{B.linecount <= 2} and \textit{B.digitProp <= 5} and \textit{135 <= B.followingSpace <= 175}
    and \textit{65 <= B.height <= 95} & Title \\
    
    10 & \textit{B.linecount <= 2} and \textit{B.digitProp <= 5} 
    and \textit{65 <= B.height <= 95} & Title \\
    
    11 & \textit{B.linecount <= 2} and \textit{B.capitalProp => 55} and \textit{595 <= B.lasthpos <= 1130} & Title \\
    
    12 & \textit{B.linecount <= 2} and \textit{B.digitProp <= 5} and \textit{B.capitalProp => 55}
    and \textit{80 <= B.followingSpace <= 135} and \textit{5849 <= B.firstvpos <= 7470} & Title \\
    
    13 & \textit{B.linecount <= 2} and \textit{B.digitProp <= 5} and \textit{B.followingSpace => 225}
    and \textit{B.height => 95} and \textit{4 <= B.wordCount <= 9} and \textit{B.precedingSpace <= 5} & Title \\
    
    14 & \textit{B.height => 95} and \textit{B.followingSpace => 225} and \textit{940 <= B.firstvpos <= 1545}
    and \textit{4 <= 4 B.wordCount <= 9} & Title \\
    
    15 & \textit{B.linecount <= 2} and \textit{B.digitProp <= 5} and \textit{B.followingSpace => 225}
    and \textit{B.height => 95} and \textit{61 <= B.wordRatio <= 70} & Title \\
    
    16 & \textit{B.linecount <= 2} and \textit{B.digitProp <= 5} and \textit{135 <= B.followingSpace <= 175}
    \\
    
    17 & \textit{B.linecount <= 2} and \textit{B.digitProp <= 5} and \textit{B.followingSpace => 225}
    and \textit{B.height => 95} and \textit{5 <= B.capitalProp <= 12.5} & Title \\
    
    \hline
    
    18 & \textit{B.lastvpos <= 900} and \textit{5 <= B.digitProp <= 80} & Header \\
    
    19 & \textit{B.lastvpos <= 900} and \textit{3015 <= B.firstHpos <= 3640}
    and \textit{B.capitalProp >= 55} & Header \\
    
    20 & \textit{B.lastvpos <= 900} and \textit{135 <= B.width <= 368.25} and \textit{B.firstvpos <= 485} & Header \\
    
    21 & \textit{B.lastvpos <= 900} and \textit{B.digitProp => 80} & Header \\
    
    22 & \textit{B.lastvpos <= 900} and \textit{12.5 <= B.capitalProp <= 55} and \textit{595 <= B.lasthpos <= 1130} \\
    
    \hline
    
    23 & \textit{B.digitProp <= 5} and \textit{B.capitalProp <= 5} and \textit{B.wordCount => 272} & Text \\
    
    24 & \textit{50 <= B.wordCount <= 272} & Text \\
    
    25 & \textit{23 <= B.wordCount <= 50} and \textit{B.capitalProp <= 5} & Text \\
    
    26 & \textit{B.digitProp <= 5} and \textit{B.capitalProp <= 5} and \textit{45 <= B.precedingSpace <= 80} & Text \\
    
    \hline

\end{tabularx}
\normalsize
\label{tab:ripperalgo-block}
\end{table}

To identify Title blocks, RIPPER produced 17 different rules whereas our rule-based system only used one. We can see that $linecount$ and $height$ are the two most frequently used features, as they occur in 11 rules. A recurring condition is that $linecount$ must be inferior to three, which is similar to the condition in our rule-based system, where a block must have no more than four lines. This confirms our intuition that a small number of lines is an important factor to detect Title blocks. 
As in our system, $followingSpace$ is also an important feature, as it appears in nine rules and must always be above a certain threshold. However, RIPPER never uses the $precedingSpace$ attribute, suggesting that only one of them is necessary to identify Title blocks. $capitalProp$ and $digitProp$ are also important features identified by RIPPER, as they respectively appear in eight and nine rules. The first one must always be above 50 \% whereas the second one must always be inferior or equal to 5 \%.

To identify Header blocks, RIPPER produced five rules, while we produced two. The condition $lastvpos <= 900$ is present in every rule, indicating that the block must be on the top part of the document. Other used features such as $digitProp$, which is always greater than 5 \%, suggest that a Header blocks  always contain numbers. To detect Headers, RIPPER relies exclusively on geometric and morphological features, unlike our system which mostly relies on semantic features such as $simHeaderset$, $simTitle$, $headerMark1$ or $headerMark2$.

Finally RIPPER produced four rules to identify Text blocks, while we produced two. The most important feature used is $wordCount$ which appears in three rules. Similarly to our system, this feature must be greater than a small threshold which is 23 words here. Unlike our system, RIPPER never uses the $linecount$ feature to detect Text blocks. Instead, it uses the $capitalProp$ and $digitProp$ features, which respectively appear in three and two rules. In every rule they appear, they must be lower or equal than 5 \%. Finally, RIPPER uses in one rule the $precedingSpace$ feature, which must be lower or equal than 80 pixels, suggesting that the space before a Text block must be small.

\subsubsection{TextLine annotation rules\\}

Table \ref{tab:ripperalgo-lines} presents all annotation rules discovered by the RIPPER algorithm for TextLine tags and their corresponding annotation labels, where $L$ is a TextLine in a document $D$ and $B$ is the TextBlock that contains $L$. Overall, RIPPER produced 19 rules whereas we produced 12, including the conflict resolution rules.

\begin{table}[H]
\small
\caption{TextLine annotation rules identified by the RIPPER algorithm}

\begin{tabularx}{\textwidth}{lXl} \hline
     \textbf{Rule} & \textbf{Condition} & \textbf{Label} \\ \hline
    1 & \textit{L.followingSpace => 75} and \textit{L.stwCapital is True} and \textit{L.blockType=title}
    and \textit{745 <= L.width <= 970} and \textit{L.vpos <= 985} & Title \\
    
    2 & \textit{L.followingSpace => 75} and \textit{L.stwCapital is True} and \textit{L.blockType=title}
    and \textit{L.capitalProp <= 5} & Title \\
    
    3 & \textit{L.followingSpace => 75} and \textit{L.stwCapital is True} and \textit{L.blockType=title}
    & Title \\
    
    4 & \textit{L.precedingSpace => 75} and \textit{L.followingSpace => 75} and \textit{335 <= L.width <= 645}
    and \textit{20 <= L.simTitle <= 29} & Title \\
    
    5 & \textit{L.followingSpace => 75} and \textit{L.stwCapital is True} and \textit{L.endsPunct is False}
    and \textit{L.capitalProp <= 5} and \textit{745 <= L.width <= 970} and \textit{L.nonAlphaProp <= 5} & Title \\
    
    6 & \textit{L.precedingSpace => 75} and \textit{L.followingSpace => 75} and \textit{L.stwCapital is True}
    and \textit{335 <= L.width <= 645} & Title \\
    
    7 & \textit{L.capitalProp => 10} and \textit{335 <= L.width <= 645} and \textit{L.height <= 35}
    and \textit{505 <= L.hpos <= 1070} and \textit{L.stwCapital is True} & Title \\
    
    8 & \textit{L.precedingSpace => 75} and \textit{L.followingSpace => 75} and \textit{745 <= L.width <= 970} & Title \\
    
    9 & \textit{L.precedingSpace => 75} and \textit{L.followingSpace => 75} and and \textit{L.stwCapital is True}
    and \textit{645 <= L.width <= 745} & Title \\ 
    \hline
    
    10 & \textit{L.capitalProp => 10} and \textit{L.vpos <= 985} and \textit{L.blockType=header} & Header \\
    
    11 & \textit{L.capitalProp => 10} and \textit{L.simHeaderSet => 85.5} and \textit{L.followingSpace => 75}
    and \textit{335 <= L.width <= 645} & Header \\
    
    12 & \textit{L.capitalProp => 10} and \textit{L.vpos <= 985} and \textit{54 <= L.simHeaderSet <= 60}
    and \textit{L.precedingSpace <= 35} & Header \\
    
    13 & \textit{L.capitalProp => 10} and \textit{L.simTitle => 42.86} and \textit{L.nonAlphaProp <= 5}
    and \textit{L.precedingSpace <= 35} and \textit{2550 <= L.hpos <= 3419} & Header \\
    
    14 &  \textit{L.capitalProp => 10} and \textit{L.vpos <= 985} and \textit{L.simTitle => 42.86}
    and \textit{L.nonAlphaProp <= 5} and \textit{335 <= L.width <= 645} & Header \\
    
    15 & \textit{505 <= L.hpos <= 1070} and \textit{L.simHeaderSet => 85.5} and \textit{335 <= L.width <= 645} & Header \\
    
    \hline
    
    16 & \textit{L.stwCapital is True} and \textit{970 <= L.width <= 1010} & Firstline \\
    
    17 & \textit{L.stwCapital is True} and \textit{L.capitalProp <= 5} & Firstline \\
    
    \hline
    
    18 & \textit{L.stwCapital is False} and \textit{L.stwDigit is False} & Text \\
    
    19 & \textit{1030 <= L.width <= 1050} & Text \\

     \hline
\end{tabularx}
\label{tab:ripperalgo-lines}
\normalsize
\end{table}

Nine rules have been produced by RIPPER to identify Title lines, where we produced four. The most important feature is $followingSpace$, as it is used in eight rules. Similarly to our system, it must always be greater than a specific threshold, here 75 pixels. $precedingSpace$ seems to be a less important feature with RIPPER than with our system, as it is only used in four rules. This would suggest that, as for Title block annotation, the space following a block is more important than the one preceding it. Our rules relied mostly on the $blockType$, $precedingSpace$ and $followingSpace$ features to identify Titles, whereas RIPPER uses these features alongside others, especially $stwCapital$ and $width$ which are both present in seven rules. $stwCapital$ is always True whereas $width$ is always smaller or equal than a specific threshold, suggesting Title lines are shorter than most lines in the document.

RIPPER produced six rules to identify Header lines. In our system, a line is labelled as Header if it is contained in a Header block. The main feature used by RIPPER is $capitalProp$ which is present in 5 rules and is always greater or equal than 10 \%. This suggests that Header lines have more capital letters than common lines. The other main features used are $vpos$, $simHeader$ and $width$. $vpos$ is always below 985 pixels, which means that the Header must be on the highest part of the document. Depending on the rules, $simHeader$ must be greater than 55 \% or 85 \% percent, suggesting the importance of the header word set. The $width$ of the line is always set in a small range between 335 and 645 pixels, suggesting Header lines are smaller than most lines of text in the document. 

Firstline annotation is the only case where RIPPER produced fewer rules than us: we produced five rules where RIPPER produced two. Similarly to our system, the main condition is that $stwCapital$ is True. The first line of a paragraph is also often indented. To detect that, our rule-based system used the $hpos$ feature whereas RIPPER uses the $width$ feature. Finally, RIPPER uses two rules to detect Text line whereas this is the default annotation in our system. In RIPPER, the main condition is that $stwCapital$ and $stwDigit$ are False or the $width$ of the line is average.

The use of the $blockType$ attribute is the main difference between our rule-based rule set and RIPPER's. In our system, the $blockType$ attribute is an important feature, as any line contained in a Header or Title block inherits this label. RIPPER also uses this feature but to a lesser extent. It is only used in three rules for Title annotation and in one rule for Header annotation, and is always used alongside other features.


\subsection{Evaluation of the RIPPER system\\}

An important question is which of the two sets of rules provides better results in terms of F1 score: RIPPER's rules or our manually designed rule set. To evaluate the rules produced by RIPPER, we have run the model through the test data set. Table \ref{tab:rippereval} shows the results of TextBlock and TextLine annotation as performed by the RIPPER algorithm.

\begin{table}[H]
\centering
\scriptsize
\caption{Precision, Recall and F1 score for TextBlock and TextLine annotation with RIPPER}

\begin{tabular}{ll|ccc|ccc|ccc|ccc}
\hline
   &  & \multicolumn{3}{c|}{\textbf{Text}} & \multicolumn{3}{c|}{\textbf{Title}} & \multicolumn{3}{c|}{\textbf{Firstline}} & \multicolumn{3}{c}{\textbf{Header}} \\
& \textbf{Cat}  & \textbf{P}       & \textbf{R}      & \textbf{F1}     & \textbf{P}       & \textbf{R}      & \textbf{F1}     & \textbf{P}       & \textbf{R}      & \textbf{F1}       & \textbf{P}       & \textbf{R}      & \textbf{F1}      \\ 
\hline

 \multirow{4}{*}{\rotatebox{90}{\textbf{TextBlock}}} & 1c   &     0.819 &  0.990 &    0.897 &     0.500 &  0.625 &    0.550  & &&& 0.857 &  0.333 &    0.480   \\
& 2c   &     0.851 &  1.000 &    0.920 &     0.500 &  0.444 &    0.471  & &&& 0.462 &  0.250 &    0.324   \\
& 3c+  &     0.852 &  0.999 &    0.920 &     0.679 &  0.297 &    0.413  & &&& 0.000 &  0.000 &    0.000   \\ 
\cline{2-14}
& Mean & 0.841 &  0.996 &    0.912 &     0.560 &  0.455 &    0.480  & &&& 0.440 &  0.194 &    0.268   \\ 

\hline

 \multirow{4}{*}{\rotatebox{90}{\textbf{TextLine}}} & 1c   &     0.871 &  0.881 &    0.876 &     0.600 &  0.290 &    0.391 &     0.644 &  0.777 &    0.704 &     0.667 &  0.053 &    0.098   \\
& 2c   &     0.923 &  0.947 &    0.935 &     0.710 &  0.268 &    0.389 &     0.754 &  0.877 &    0.811 &     0.667 &  0.065 &    0.118   \\
& 3c+  &     0.876 &  0.908 &    0.892 &     0.610 &  0.207 &    0.309 &     0.614 &  0.709 &    0.658 &     0.000 &  0.000 &    0.000   \\ 
\cline{2-14}
& Mean & 0.890 &  0.912 &    0.901 &     0.640 &  0.255 &    0.363 &     0.671 &  0.788 &    0.724 &     0.444 &  0.039 &    0.072   \\
\hline
\end{tabular}
\normalsize
\label{tab:rippereval}

\end{table}

TextBlock annotation performs the best on the 1c layout category. The annotation of Text elements provides the best results with an F1 score of 0.912 on average. Title and Header annotations perform much worse with F1 scores of 0.48 and 0.26 respectively. The main cause of error is a confusion between Header and Title blocks. Every Header block mislabelled as Title contained fewer than three lines, and vice versa. This suggests that more conditions are required to distinguish between small Header blocks and Titles. Finally, many Header blocks were mislabelled as Text, suggesting a lack of rules to detect them. 

TextLine annotation performs the best on the 2c layout category. Here again, the annotation of Text elements provides the best results with an F1 score of 0.901 on average. Firstline annotation comes second with an average F1 score of 0.724. Title and Header annotations are also disappointing, with respectively 0.36 and 0.07 of F1 scores on average. Like in TextBlock annotation, many Title lines were mislabelled as Header. Most Title and Header lines were mislabelled as Text because of their width. This suggests either that the rules to detect these two labels are not precise enough, or that there are not enough rules in RIPPER's rule set. There is a confusion between Text lines and Firstline: 519 of Text lines were incorrectly labelled as Firstline because they started with a capital letter. A similar amount of Firstline was mislabelled as Text because of a low-quality OCR.

Overall, we can observe that RIPPER's TextLine annotation obtains lower F1 scores than its TextBlock annotation. This result is on the opposite of the evaluation of our rule-based system. TextBlock annotation is designed as an intermediary step that should facilitate the TextLine annotation, which is the final result of the algorithm. Unlike our rule-based system however, RIPPER uses only sparsely the TextBlock annotations to obtain the TextLine annotations. This would explain its poor final results on TextLine annotation.


\subsection{Evaluation of the Gradient Boosting algorithms\\}

Table \ref{tab:mlevaluation} shows the results for TextBlock and TextLine annotation performed by the Gradient Boosting algorithms on the test set.

\begin{table}[H]
\centering
\scriptsize
\caption{Precision, Recall and F1 score for TextBlock and TextLine annotation with Gradient Boosting}

\begin{tabular}{ll|ccc|ccc|ccc|ccc}
\hline
   &  & \multicolumn{3}{c|}{\textbf{Text}} & \multicolumn{3}{c|}{\textbf{Title}} & \multicolumn{3}{c|}{\textbf{Firstline}} & \multicolumn{3}{c}{\textbf{Header}} \\
& \textbf{Cat}  & \textbf{P}       & \textbf{R}      & \textbf{F1}     & \textbf{P}       & \textbf{R}      & \textbf{F1}     & \textbf{P}       & \textbf{R}      & \textbf{F1}       & \textbf{P}       & \textbf{R}      & \textbf{F1}      \\
\hline

 \multirow{4}{*}{\rotatebox{90}{\textbf{TextBlock}}} & 1c   &     0.863 &  0.962 &    0.910 &     0.500 &  0.375 &    0.429  & &&& 0.667 &  0.222 &    0.333   \\
& 2c   &     0.891 &  0.989 &    0.937 &     0.706 &  0.667 &    0.686  & &&& 0.625 &  0.208 &    0.312   \\
& 3c+  &     0.967 &  0.863 &    0.912 &     0.789 &  0.234 &    0.361  & &&& 0.000 &  0.000 &    0.000   \\ 
\cline{2-14}
& Mean & 0.907 &  0.938 &    0.920 &     0.665 &  0.425 &    0.492  & &&& 0.431 &  0.144 &    0.215   \\ 

\hline

 \multirow{4}{*}{\rotatebox{90}{\textbf{TextLine}}} & 1c   &     0.778 &  0.983 &    0.868 &     0.667 &  0.194 &    0.300 &     0.641 &  0.207 &    0.312 &     1.000 &  0.211 &    0.348 \\
& 2c   &     0.905 &  0.949 &    0.926 &     0.758 &  0.305 &    0.435 &     0.727 & 0.756 &    0.741 &     0.875 &  0.113 &    0.200 \\
& 3c+  &     0.856 &  0.950 &    0.901 &     0.941 &  0.264 &    0.413 &     0.667 &  0.483 &    0.560 &     0.000 &  0.000 &    0.000 \\
\cline{2-14}
& Mean & 0.846 &  0.961 &    0.898 &     0.788 &  0.254 &    0.383 &     0.678 &  0.482 &    0.538 &     0.625 &  0.108 &    0.183 \\
\hline
\end{tabular}
\normalsize
\label{tab:mlevaluation}

\end{table}

For TextBlock annotation, the model performs best on the 2c category. The annotation of Text elements provides the best results with a F1-score of 0.92 on average. The performance for Title annotation is average with a mean F1-score of 0.49, but with a mean Precision of 0.66. Finally, Header annotation performs the worst with a mean F1 score of 0.21 and an average Precision of 0.43. Most errors are Header blocks mislabelled as Title (23 \%) and Text blocks mislabelled as Header (17 \%). Most mislabelled Title and Header blocks were labelled as Text, suggesting the model only covers a few instances of both labels.

For TextLine annotation, the model also performs the best on the 2c category, except for Header annotation where it performs the best on the 1c category. As for TextBlock annotation, Text lines annotation has the best performance with a mean F1 score of 0.89 and a Recall of 0.96. Firstline annotation is average, with a mean F1 score of 0.53 and an average Precision of 0.67. Both Header and Title annotation have bad performances with a mean F1 score of 0.18 and 0.38, but an average Precision score of 0.62 and 0.78 respectively. Every mislabelled Text line was labelled as Firstline. On the other hand, most mislabelled Firstline, Title and Header lines were labelled as Text, indicating the model only covers a few instances of these labels, as for TextBlock annotation.

\subsection{Final comparison between the models\\}
\label{modelcomparison}

\begin{figure}[htbp]
    \centering
        \caption{Mean F1 score of the three models for TextBlock (left) and TextLine (right) annotations}
        
    \subfigure{\label{fig:ml_comparison_block}\includegraphics[width=75mm]{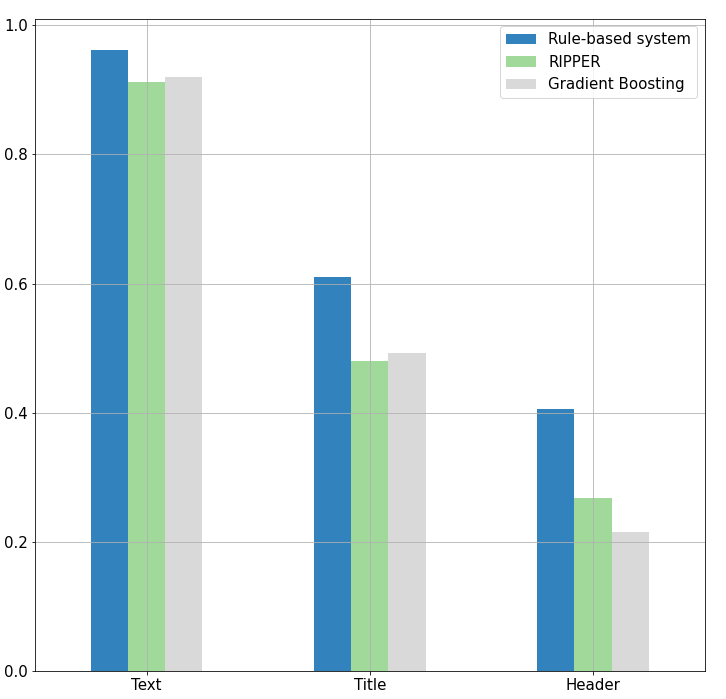}}
    \subfigure{\label{fig:ml_comparison_lines}\includegraphics[width=75mm]{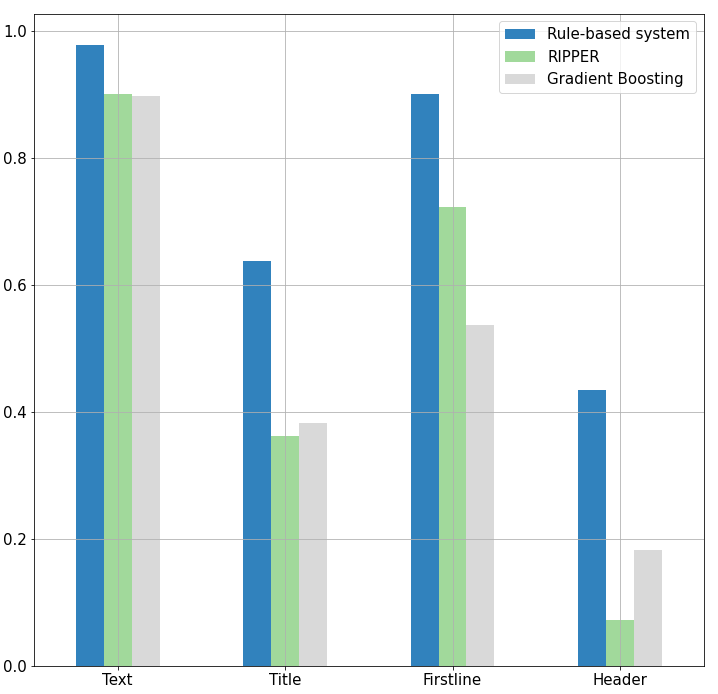}}



    \label{fig:f1meangraph}
\end{figure}

Table \ref{tab:comparaisoneval} presents the mean Precision, Recall and F1 scores of the three systems for TextBlock and TextLine annotation on the test set: the rule-based system, the RIPPER algorithm and the Gradient Boosting algorithm. Figure \ref{fig:f1meangraph} shows the average F1 scores of the three models for TextBlock and TextLine annotations.

\begin{table}[htbp]
\centering
\caption{Mean Precision, Recall and F1 scores of the three systems on the TextBlock and TextLine annotation tasks}
\scriptsize
\begin{tabular}{ll|ccc|ccc|ccc|ccc}
\hline
   &  & \multicolumn{3}{c|}{\textbf{Text}} & \multicolumn{3}{c|}{\textbf{Title}} & \multicolumn{3}{c|}{\textbf{Firstline}} & \multicolumn{3}{c}{\textbf{Header}} \\
& \textbf{Cat}  & \textbf{P}       & \textbf{R}      & \textbf{F1}     & \textbf{P}       & \textbf{R}      & \textbf{F1}     & \textbf{P}       & \textbf{R}      & \textbf{F1}       & \textbf{P}       & \textbf{R}      & \textbf{F1}      \\
\hline

 \multirow{4}{*}{\rotatebox{90}{\textbf{TextBlock}}} 
& Rule-based & \textbf{0.959}  & 0.966  & \textbf{0.962}  & 0.600   & \textbf{0.639}   & \textbf{0.610}  & &&& \textbf{0.726}   & \textbf{0.298}   & \textbf{0.406}   \\ 
& RIPPER & 0.841 &  \textbf{0.996} &    0.912 &     0.560 &  0.455 &    0.480  & &&& 0.440 &  0.194 &    0.268   \\ 
& Gradient Boosting & 0.907 &  0.938 &    0.920 &     \textbf{0.665} &  0.425 &    0.492  & &&& 0.431 &  0.144 &    0.215   \\
&&&&&&&&&&&&&\\
\hline

 \multirow{4}{*}{\rotatebox{90}{\textbf{TextLine}}} 
& Rule-based & \textbf{0.969}   & \textbf{0.991}  & \textbf{0.979}  & 0.595   & \textbf{0.733}   & \textbf{0.639}  & \textbf{0.949} & \textbf{0.861}    & \textbf{0.902}   & \textbf{0.803}   & \textbf{0.348}   & \textbf{0.435}   \\
& RIPPER & 0.890 &  0.912 &    0.901 &     0.640 &  0.255 &    0.363 &     0.671 &  0.788 &    0.724 &     0.444 &  0.039 &    0.072   \\
& Gradient Boosting & 0.846 &  0.961 &    0.898 &     \textbf{0.788} &  0.254 &    0.383 &     0.678 &  0.482 &    0.538 &     0.625 &  0.108 &    0.183 \\
&&&&&&&&&&&&&\\
\hline

\end{tabular}
\normalsize
\label{tab:comparaisoneval}

\end{table}

All three models have good performances for Text blocks annotation, with the minimum reaching an average F1 score of 0.91. Our rule-based system has the best Precision and F1 score, whereas RIPPER has the best Recall score. The performances for Title block annotation are average for all three models. Gradient Boosting has the best mean Precision score with 0.66, while our rule-based system has the best mean Recall and F1 score. Furthermore, our system seems more stable, as Precision, Recall and F1 are nearly equal, unlike the other two models. Finally, the performances for Header block annotation go from bad to good. Every system has a much better Precision score than a Recall, suggesting that defining exhaustive rules to detect Header is a difficult task. However, our system has the best performances in Precision, Recall and F1 score once again.

Similarly to Text blocks annotation, all three models have very good performances for Text line annotation. Our rule-based system has the best results in each three metrics. The performances for Title line annotation go from bad to very good. Like Title block annotation, Gradient Boosting has the best mean Precision score. However, our system has the best mean Recall and F1 score by far. The performances for Firstline annotation go from average to very good. Our system has the best scores in every metric, followed by the RIPPER system. Surprisingly, Gradient Boosting only reaches an average F1 score of 0.53, suggesting that paragraphs are easier to detect with simple rules. Finally, the performance for Header line annotation go from very bad to very good. Similarly to TextBlock annotation, every system has a much better Precision than Recall. In this case as well, our system reaches the best scores in each metric. 

Our rule-based system outperforms the two other models in nearly all evaluations. It has especially better Recall results, indicating that our system covers more types of every logical label than the other two models. When comparing RIPPER with Gradient Boosting, we can observe that Gradient Boosting has better Precision scores but RIPPER has better Recall scores.

Despite its disappointing performances, RIPPER can produce very precise rules which are sometimes better than our manually crafted rules. Furthermore, RIPPER has better Precision than Recall scores on average, suggesting this algorithm is better at producing fine-grained rules than general ones. As producing rules by hand is a time-consuming task, it would be interesting in future works to first use RIPPER in an exploratory manner to produce a base rule set with high Precision scores. This rule set could then be manually updated in order to improve the performances of the rules. 

Similarly, Gradient Boosting can reach very high Precision scores, as seen with Title lines annotation. Thus, it would be interesting to produce a hybrid system which would  either use rules or Machine Learning algorithms to identify a specific logical label.

\section{Conclusion\\}
\label{conclusion}

In this article, we have compared the performances of three systems for Logical Layout Analysis applied on XML ALTO: a rule-based system, the RIPPER Rule Learning algorithm and Gradient Boosting. All three of them are used in the same general pipeline: a set of features is first extracted from the document, and the system is then used to assign logical labels to TextBlock and TextLine elements. 

The comparison between the performances of the three systems shows that our rule-based system outperforms the two other models in nearly all evaluations. Its higher Recall scores suggest that this system covers more types of every logical label than the other two models. The evaluation also confirms that our system can be used to produce annotated data sets that are large enough to envisage Machine Learning or deep learning approaches. However, both RIPPER and Gradient Boosting can reach very high Precision scores. Combining rules and Machine Learning models into hybrid systems could potentially  provide even better performances. Thus, we plan in future works to produce such hybrid systems and evaluate them.

As stated earlier, the layout in historical documents evolves rapidly, especially in newspapers. It is then necessary to develop systems dedicated to a specific publication period. Although RIPPER's performances are disappointing, it is a valuable tool to explore a data set and quickly create a rule set, which can then be updated manually. As such, we plan in future works to use Rule Learning algorithms such as RIPPER to help creating rule sets adapted to specific publication periods. 

\section*{Competing Interest\\}
The authors declare that they have no competing interests.

\section*{List of abbreviations\\}
\begin{description}
    \item[OCR :] Optical Character Recognition
    \item[PLA :] Physical Layout Analysis
    \item[LLA :] Logical Layout Analysis
    \item[HAC :] Hierarchical Agglomerative Clustering
    \item[CNN :] Convolutional Neural Network
    \item[LSTM :] Long Short-Term Memory
    \item[MARG :] Medical Articles Record Groundtruth
    \item[CRF :] Conditional Random Field
    \item[RIPPER :] Repeated Incremental Pruning to Produce Error Reduction
    \item[IREP :] Incremental Reduced Error Pruning
    \item[FOIL :] First Order Inductive Learner
    \item[DL :] Description Length
    \item[SVM :] Support Vector Machine
\end{description}

\section*{Authors' contributions\\}

All authors contributed equally to the study conception and design. Nicolas Gutehrlé carried out the background description, the creation and annotation of the data set, the construction of the rule-based system, the training of the Rule-Learning and Machine-Learning algorithm, the evaluation of the three systems and drafted the manuscript. Iana Atanassova supervised all of the above tasks and participated in improving the manuscript and the presentation of the results. All authors read and approved the final manuscript.

\section*{Author's information\\}

Nicolas Gutehrlé is currently a PhD Student in Natural Language Processing at the CRIT laboratory at Université de Bourgogne Franche-Comté, under the supervision of Dr. Iana Atanassova.

Dr. Iana Atanassova is associate professor in Natural Language Processing at the CRIT laboratory at Université de Bourgogne Franche-Comté and at the Institut Universitaire de France (IUF). She supervises the EMONTAL project (Extraction and Ontology Modeling of Subjects and Places for the Exploitation of the Documentary Funds of Bourgogne Franche-Comté, 2020--2023) funded by the Région Bourgogne Franche-Comté, France.

\section*{Acknowledgments\\}

This research is supported by the Région Bourgogne Franche-Comté, France, as part of the EMONTAL project (Extraction and Ontology Modeling of Subjects and Places for the Exploitation of the Documentary Funds of Bourgogne Franche-Comté, 2020--2023).

\bibliographystyle{plainnat}
\bibliography{bibliography}

\begin{thebibliography}{30}
\providecommand{\natexlab}[1]{#1}
\providecommand{\url}[1]{\texttt{#1}}
\expandafter\ifx\csname urlstyle\endcsname\relax
  \providecommand{\doi}[1]{doi: #1}\else
  \providecommand{\doi}{doi: \begingroup \urlstyle{rm}\Url}\fi

\bibitem[Akbik et~al.(2018)Akbik, Blythe, and Vollgraf]{akbik2018coling}
Alan Akbik, Duncan Blythe, and Roland Vollgraf.
\newblock Contextual string embeddings for sequence labeling.
\newblock In \emph{{COLING} 2018, 27th International Conference on
  Computational Linguistics}, pages 1638--1649, 2018.

\bibitem[Akl et~al.(2019)Akl, Gupta, and Mariko]{akl-etal-2019-fintoc}
Hanna~Abi Akl, Anubhav Gupta, and Dominique Mariko.
\newblock {F}in{TOC}-2019 shared task: Finding title in text blocks.
\newblock In \emph{Proceedings of the Second Financial Narrative Processing
  Workshop (FNP 2019)}, pages 58--62, Turku, Finland, September 2019.
  Link{\"o}ping University Electronic Press.
\newblock URL \url{https://www.aclweb.org/anthology/W19-6408}.

\bibitem[Ayatollahi and Nafchi(2013)]{phibd}
S.M. Ayatollahi and Hossein Nafchi.
\newblock Persian heritage image binarization competition (phibc 2012).
\newblock pages 1--4, 03 2013.
\newblock ISBN 978-1-4673-6204-7.
\newblock \doi{10.1109/PRIA.2013.6528442}.

\bibitem[Barman et~al.(2020)Barman, Ehrmann, Clematide, Oliveira, and
  Kaplan]{Barman2020CombiningVA}
Rapha{\"e}l Barman, Maud Ehrmann, S.~Clematide, S.~Oliveira, and F.~Kaplan.
\newblock Combining visual and textual features for semantic segmentation of
  historical newspapers.
\newblock \emph{ArXiv}, abs/2002.06144, 2020.

\bibitem[Bojanowski et~al.(2016)Bojanowski, Grave, Joulin, and
  Mikolov]{bojanowski2016enriching}
Piotr Bojanowski, Edouard Grave, Armand Joulin, and Tomas Mikolov.
\newblock Enriching word vectors with subword information.
\newblock \emph{arXiv preprint arXiv:1607.04606}, 2016.

\bibitem[Bulacu et~al.(2007)Bulacu, van Koert, Schomaker, and van~der
  Zant]{4378732}
Marius Bulacu, Rutger van Koert, Lambert Schomaker, and Tijn van~der Zant.
\newblock Layout analysis of handwritten historical documents for searching the
  archive of the cabinet of the dutch queen.
\newblock In \emph{Ninth International Conference on Document Analysis and
  Recognition (ICDAR 2007)}, volume~1, pages 357--361, 2007.
\newblock \doi{10.1109/ICDAR.2007.4378732}.

\bibitem[Chen and Seuret(2017)]{chen2017convolutional}
Kai Chen and Mathias Seuret.
\newblock Convolutional neural networks for page segmentation of historical
  document images, 2017.

\bibitem[Clausner et~al.(2015)Clausner, Papadopoulos, Pletschacher, and
  Antonacopoulos]{7333898}
Christian Clausner, Christos Papadopoulos, Stefan Pletschacher, and Apostolos
  Antonacopoulos.
\newblock The enp image and ground truth dataset of historical newspapers.
\newblock In \emph{2015 13th International Conference on Document Analysis and
  Recognition (ICDAR)}, pages 931--935, 2015.
\newblock \doi{10.1109/ICDAR.2015.7333898}.

\bibitem[Cohen(1995)]{cohen1995repeated}
William~W Cohen.
\newblock Repeated incremental pruning to produce error reduction.
\newblock In \emph{Machine Learning Proceedings of the Twelfth International
  Conference ML95}, 1995.

\bibitem[F{\"u}rnkranz and Widmer(1994)]{furnkranz1994incremental}
Johannes F{\"u}rnkranz and Gerhard Widmer.
\newblock Incremental reduced error pruning.
\newblock In \emph{Machine Learning Proceedings 1994}, pages 70--77. Elsevier,
  1994.

\bibitem[Gutehrlé and Atanassova(2021)]{nicolas_gutehrle_2021_5752440}
Nicolas Gutehrlé and Iana Atanassova.
\newblock {Dataset for Logical-layout analysis on French historical
  newspapers}, October 2021.

\bibitem[Hearst(1997)]{texttiling}
Marti~A. Hearst.
\newblock Texttiling: Segmenting text into multi-paragraph subtopic passages.
\newblock \emph{Comput. Linguist.}, 23\penalty0 (1):\penalty0 33–64, March
  1997.
\newblock ISSN 0891-2017.

\bibitem[Hébert et~al.(2014)Hébert, Palfray, Nicolas, Tranouez, and
  Paquet]{crfarticle}
David Hébert, Thomas Palfray, Stéphane Nicolas, Pierrick Tranouez, and
  Thierry Paquet.
\newblock Automatic article extraction in old newspapers digitized collections.
\newblock \emph{ACM International Conference Proceeding Series}, 05 2014.
\newblock \doi{10.1145/2595188.2595195}.

\bibitem[Kise et~al.(1999)Kise, Iwata, and Matsumoto]{Kise1999OnTA}
K.~Kise, M.~Iwata, and Keinosuke Matsumoto.
\newblock On the application of voronoi diagrams to page segmentation.
\newblock 1999.

\bibitem[Klampfl and Kern(2013)]{Klampfl2013AnUM}
S.~Klampfl and Roman Kern.
\newblock An unsupervised machine learning approach to body text and table of
  contents extraction from digital scientific articles.
\newblock In \emph{TPDL}, 2013.

\bibitem[Nagy et~al.(1992)Nagy, Seth, and Viswanathan]{Nagy1992APD}
G.~Nagy, S.~Seth, and M.~Viswanathan.
\newblock A prototype document image analysis system for technical journals.
\newblock \emph{Computer}, 25:\penalty0 10--22, 1992.

\bibitem[Namboodiri and Jain(2007)]{docstructure}
Anoop Namboodiri and Anil Jain.
\newblock \emph{Document Structure and Layout Analysis}, pages 29--48.
\newblock 03 2007.
\newblock ISBN 978-1-84628-501-1.
\newblock \doi{10.1007/978-1-84628-726-8_2}.

\bibitem[Niyogi and Srihari(1995)]{delos}
D.~Niyogi and S.N. Srihari.
\newblock Knowledge-based derivation of document logical structure.
\newblock In \emph{Proceedings of 3rd International Conference on Document
  Analysis and Recognition}, volume~1, pages 472--475 vol.1, 1995.
\newblock \doi{10.1109/ICDAR.1995.599038}.

\bibitem[O'Gorman(1993)]{OGorman1993TheDS}
L.~O'Gorman.
\newblock The document spectrum for page layout analysis.
\newblock \emph{IEEE Trans. Pattern Anal. Mach. Intell.}, 15:\penalty0
  1162--1173, 1993.

\bibitem[Pedregosa et~al.(2011)Pedregosa, Varoquaux, Gramfort, Michel, Thirion,
  Grisel, Blondel, Prettenhofer, Weiss, Dubourg, Vanderplas, Passos,
  Cournapeau, Brucher, Perrot, and Duchesnay]{scikit-learn}
F.~Pedregosa, G.~Varoquaux, A.~Gramfort, V.~Michel, B.~Thirion, O.~Grisel,
  M.~Blondel, P.~Prettenhofer, R.~Weiss, V.~Dubourg, J.~Vanderplas, A.~Passos,
  D.~Cournapeau, M.~Brucher, M.~Perrot, and E.~Duchesnay.
\newblock Scikit-learn: Machine learning in {P}ython.
\newblock \emph{Journal of Machine Learning Research}, 12:\penalty0 2825--2830,
  2011.

\bibitem[Pennington et~al.(2014)Pennington, Socher, and
  Manning]{pennington-etal-2014-glove}
Jeffrey Pennington, Richard Socher, and Christopher Manning.
\newblock {G}lo{V}e: Global vectors for word representation.
\newblock In \emph{Proceedings of the 2014 Conference on Empirical Methods in
  Natural Language Processing ({EMNLP})}, pages 1532--1543, Doha, Qatar,
  October 2014. Association for Computational Linguistics.
\newblock \doi{10.3115/v1/D14-1162}.
\newblock URL \url{https://aclanthology.org/D14-1162}.

\bibitem[Quinlan(2005)]{Quinlan2005LearningLD}
J.~Ross Quinlan.
\newblock Learning logical definitions from relations.
\newblock \emph{Machine Learning}, 5:\penalty0 239--266, 2005.

\bibitem[Ramakrishnan et~al.(2012)Ramakrishnan, Patnia, Hovy, and
  Burns]{lapdftext}
Cartic Ramakrishnan, Abhishek Patnia, Eduard Hovy, and Gully Burns.
\newblock Layout-aware text extraction from full-text pdf of scientific
  articles.
\newblock \emph{Source code for biology and medicine}, 7:\penalty0 7, 05 2012.
\newblock \doi{10.1186/1751-0473-7-7}.

\bibitem[Riedl et~al.(2019)Riedl, Betz, and
  Pad{\'o}]{riedl-etal-2019-clustering}
Martin Riedl, Daniela Betz, and Sebastian Pad{\'o}.
\newblock Clustering-based article identification in historical newspapers.
\newblock In \emph{Proceedings of the 3rd Joint {SIGHUM} Workshop on
  Computational Linguistics for Cultural Heritage, Social Sciences, Humanities
  and Literature}, pages 12--17, Minneapolis, USA, June 2019. Association for
  Computational Linguistics.
\newblock \doi{10.18653/v1/W19-2502}.

\bibitem[Sammut and Webb(2017)]{sammut-encyclopedia}
Claude Sammut and Geoffrey~I. Webb, editors.
\newblock \emph{Encyclopedia of Machine Learning and Data Mining}.
\newblock Springer, 2017.
\newblock ISBN 978-1-4899-7685-7.
\newblock \doi{10.1007/978-1-4899-7687-1}.

\bibitem[Shen et~al.(2020)Shen, Zhang, and Dell]{Shen2020ALD}
Zejiang Shen, Kaixuan Zhang, and Melissa Dell.
\newblock A large dataset of historical japanese documents with complex
  layouts.
\newblock \emph{2020 IEEE/CVF Conference on Computer Vision and Pattern
  Recognition Workshops (CVPRW)}, pages 2336--2343, 2020.

\bibitem[Simistira et~al.(2016)Simistira, Seuret, Eichenberger, Garz, Liwicki,
  and Ingold]{Simistira2016DIVAHisDBAP}
Fotini Simistira, Mathias Seuret, Nicole Eichenberger, A.~Garz, M.~Liwicki, and
  R.~Ingold.
\newblock Diva-hisdb: A precisely annotated large dataset of challenging
  medieval manuscripts.
\newblock \emph{2016 15th International Conference on Frontiers in Handwriting
  Recognition (ICFHR)}, pages 471--476, 2016.

\bibitem[Tibbo(2007)]{Tibbo2007PrimarilyHI}
H.~Tibbo.
\newblock Primarily history in america: How u.s. historians search for primary
  materials at the dawn of the digital age.
\newblock \emph{American Archivist}, 66:\penalty0 9--50, 2007.

\bibitem[Zhong et~al.(2019)Zhong, Tang, and Jimeno-Yepes]{Zhong2019PubLayNetLD}
Xu~Zhong, J.~Tang, and Antonio Jimeno-Yepes.
\newblock Publaynet: Largest dataset ever for document layout analysis.
\newblock \emph{2019 International Conference on Document Analysis and
  Recognition (ICDAR)}, pages 1015--1022, 2019.

\bibitem[Zulfiqar et~al.(2019)Zulfiqar, Ul-Hasan, and Shafait]{8946046}
Annus Zulfiqar, Adnan Ul-Hasan, and Faisal Shafait.
\newblock Logical layout analysis using deep learning.
\newblock In \emph{2019 Digital Image Computing: Techniques and Applications
  (DICTA)}, pages 1--5, 2019.
\newblock \doi{10.1109/DICTA47822.2019.8946046}.

\end{thebibliography}


\end{document}